**Title:** An AI-Based Behavioral Health Safety Filter and Dataset for Identifying Mental Health Crises in Text-Based Conversations

**Authors:** Benjamin W. Nelson, Ph.D.[1,2], Celeste Wong, MPH[1], Matthew T. Silvestrini, Ph.D.[1], Sooyoon Shin, Ph.D.[1], Alanna Robinson, MSW, LICSW[1], Jessica Lee, MS, MBA[1], Eric Yang, MBI[1], John Torous, MD[2], Andrew Trister, MD PhD, MSE[1]

**Affiliations:**

[1]Verily Life Sciences, South San Francisco, CA, United States

[2]Division of Digital Psychiatry, Department of Psychiatry, Harvard Medical School and Beth Israel Deaconess Medical Center, Boston, MA

**Corresponding Author:**

Benjamin W. Nelson, PhD

Verily Life Sciences

269 E Grand Ave, South San Francisco, CA 94080

Email: bwn@verily.com

**Abstract**

**Background**

Large language models often mishandle psychiatric emergencies, offering harmful or inappropriate advice and enabling destructive behaviors.

**Methods**

This study evaluated the Verily behavioral health safety filter (VBHSF) on two datasets: the Verily Mental Health Crisis Dataset v1.0 containing 1,800 simulated messages and the NVIDIA Aegis AI Content Safety Dataset subsetted to 794 mental health-related messages. The two datasets were clinician-labelled and we evaluated performance using the clinician labels as reference. Additionally, we carried out comparative performance analyses (clinician labels as reference) against two open source, content moderation guardrails: OpenAI Omni Moderation Latest and NVIDIA NeMo Guardrails.

**Results**

The VBHSF demonstrated, well-balanced performance on the Verily Mental Health Crisis Dataset v1.0, achieving high sensitivity (0.990) and specificity (0.992) in detecting any mental health crises. It achieved an F1-score of 0.939, sensitivity ranged from 0.917-0.992, and specificity was $\geq 0.978$ in identifying specific crisis categories. When evaluated against the NVIDIA Aegis AI Content Safety Dataset 2.0, VBHSF performance remained highly sensitive (0.982) and accuracy (0.921) with reduced specificity (0.859). When compared with the NVIDIA NeMo and OpenAI Omni Moderation Latest guardrails, the VBHSF demonstrated superior performance metrics across both datasets, achieving significantly higher sensitivity in all cases (all $p < 0.001$) and higher specificity relative to NVIDIA NeMo ($p < 0.001$), but not to OpenAI Omni Moderation Latest ($p = 0.094$). NVIDIA NeMo and OpenAI Omni Moderation Latest exhibited inconsistent performance across specific crisis types, with sensitivity for some categories falling below 0.10.

**Conclusion**

Overall, the VBHSF demonstrated robust, generalizable performance that prioritizes sensitivity to minimize missed crises, a crucial feature for healthcare applications.

## Introduction

Psychiatric emergencies are prevalent, multidimensional, and rising[1,2]. Detecting crises is particularly challenging, as clinicians perform only slightly better than chance[3,4] and individuals often communicate distress indirectly[5–7]. People experiencing psychiatric challenges are increasingly turning to chatbots leveraging large language models (LLMs) and generative artificial intelligence (AI) for information and emotional support[8,9].

LLMs have been implicated in high-profile clinical failures, including missed warning signs, unsafe advice, or even harm facilitation. These failures include enabling suicide[10–12], promoting violence[13], and providing dangerous guidance on self-harm[14], drug use, and eating disorders[15,16]. These incidents have drawn national attention, prompting American Psychological Association testimony before the US Senate Judiciary Committee[17], a bipartisan letter by 54 state attorney generals on AI and exploitation of children[18] and a special FDA panel planned for mental health and AI in November 2025[19].

Accurate detection of psychiatric crises is critical for ensuring safety, yet significant challenges remain. Indicators of risk are often nuanced and subtle, and multidimensional spanning suicide to abuse and requiring clinicians to recognize signals[5–7]. While developers have implemented general safety filters[20–26], current solutions are designed for generic risks (e.g., sexual content, general violence)[21,23] and are not fit-for-purpose for detecting psychiatric crises.

Several critical gaps remain before LLMs can reliably detect and respond to psychiatric crises. Common frameworks and evaluation datasets are needed to define crises and evaluate model performance[27]. Existing benchmark datasets lack mental health messages, and/or clinical annotation, resulting in low-quality datasets for developing behavioral health safety filters[28–30]. Consequently, there are two key challenges for developing such filters: general-purpose safety filters may lack clinical nuance for psychiatric crisis detection, and available datasets are rarely clinically validated or specific to psychiatric risk.

To address these challenges, we 1) identified eight dimensions of mental health crises that represent the most urgent and high-risk presentations in collaboration with clinical experts: abuse, neglect, eating-disorder behaviors, psychosis, self-harm, suicide, substance misuse, violence towards others, and mixed presentations corresponding to risk factors that clinicians are trained to monitor closely and, in many cases, are legally mandated to report; 2) developed the Verily behavioral health safety filter (VBHSF) to identify these crises in user messages using a simulated dataset of 1,800 crisis and non-crisis messages reflecting real-world texting behaviors that were labeled and reviewed by two licensed clinicians; 3) evaluated the performance of VBHSF in the Verily Mental Health Crisis Dataset v1.0 (internal evaluation) and in a subset of mental health messages from the NVIDIA Aegis AI Content Safety Dataset 2.0 (external evaluation); and 4) compared the VBHSF performance to two state-of-the-art general-purpose guardrails.

## Methods

### Verily Mental Health Crisis Dataset v1.0.

We developed a simulated dataset of 1,800 messages designed to reflect realistic digital communication patterns across mental health crisis and non-crisis messages, herein referred to as the Verily dataset. We used pre-trained language models to synthesize the dataset to mitigate the unavailability of highly sensitive mental health crisis messages as has been done with other datasets[21,22]. The crisis messages (n = 900) included 100 messages for each of the following categories: abuse, neglect, eating-disorder behaviors, psychosis, self-harm, suicide, substance misuse, violence toward others, and mixed crisis presentations categories. These categories reflect clinical understanding of risks for psychiatric emergencies and adverse outcomes[31–34], often serious enough to require mandated reporting. The non-crisis messages (n = 900) consisted of health-related conversational messages including, appointment logistics, behavior change, follow-up care, follow-up clarification, health education, health monitoring, insurance and billing, lab and wearable data interpretation, lifestyle modification, locating providers, medical record request, preventative care, referral request, test scheduling, symptom tracking, medication pickup logistics, chronic condition management, and diet and nutrition.

Crisis messages consisted of three categories, with 91.10% being classified as direct explicit expression of risk (e.g.,*"I want to kill myself"*), 8.90% classified as ambiguous statements of harm (e.g., *"I feel like I'm always walking on eggshells when he's around n it scares me"*), and 8.6% including slang or masked language (e.g., *"13'ing"* for suicide or *"relief lines"* for self-harm scars). Direct and ambiguous expressions of risk were mutually exclusive, but either categories could overlap with masked language, as in *"dude teh urge to unalive myself is getting stronger"*, where *"unalive"* serves as both a clear indicator of harm and slang for suicide. These categories were constructed to reflect a broad spectrum of real-world language mechanics and texting behaviors from the literature[35–38] and online communities[39–41]. The full dataset contained the following language characteristics that were not mutually exclusive:

- **Language Mechanics Errors (Overall 55.90%, Crisis 56.60%, Non-crisis 55.30%):** Spelling, grammar, punctuation, capitalization, or spacing issues (e.g., *"definately"*, *"recieve"*, *"seperate"*).
- **Textese (Overall 45.80%, Crisis 42.90%, Non-crisis 48.70%):** Deliberate shorthand or abbreviations (e.g., *"u"*, *"bc"*).
- **Emojis and symbolic markers (Overall 13.50%, Crisis 15.70%, Non-crisis 11.30%):**

Symbols conveying affect (e.g., 😢 🔪 🩸).

Two licensed clinicians (BWN, AR) independently reviewed and annotated 1,800 messages as crisis or non-crisis with high inter-rater reliability (Cohen's κ = 0.99). Discrepancies were resolved through adjudication and consensus.

**NVIDIA Aegis AI Content Safety Dataset 2.0.**

We utilized a subset of the NVIDIA Aegis AI Content Safety Dataset 2.0[29] (cc-by-4.0), specifically 794 messages (397 crisis, 397 non-crisis) that were human data, as the NVIDIA dataset, focusing on "Suicide and Self Harm," excluding samples from the Suicide Detection dataset and duplicate rows. Two licensed clinicians (BWN, JT) independently reviewed messages, reclassifying 6.927% of the total (26 "safe" to "unsafe," 29 "unsafe" to "safe") to ensure label accuracy.

**Safety Filters, Content Moderation, and Guardrail Models.**

**Verily behavioral health safety filter (VBHSF).** The VBHSF is a transformer-based LLM (GPT architecture) that uses advanced prompt engineering and clinical reasoning to detect the presence of a crisis within a message and classify the specific crisis type.

**Open Source Safety Guardrails.** We selected two open-source content moderation guardrails, with specific categories that included safety and mental health. The first, OpenAI's omni-moderation-latest model[21,24], built on GPT-4o, automatically flags text and images for policy violations across 13 predefined categories and returns a binary flag and JSON output. The second, nvidia/llama-3.1-nemoguard-8b-content-safety[20,23] is an open-source model that classifies content across 23 risk categories and outputs a JSON.

**Statistical Analysis**

All statistical analyses were conducted using Python (version 3.9.6). The analyses were structured to evaluate the internal performance of the VBHSF, followed by evaluation with the NVIDIA dataset, and against external guardrails.

The internal evaluation assessed the VBHSF on the Verily dataset. This included the performance of the Stage 1 classifier for overall crisis detection, and the performance of the Stage 2 multi-label classifier on messages flagged as "Crisis" by the Stage 1 model. The Stage 2 evaluation used a one-vs-rest framework, with outcomes including per-category metrics, a confusion matrix to analyze misclassification patterns, and macro-averaged metrics to assess balanced performance across categories. On the NVIDIA dataset, only the performance of the Stage 1 classifier for overall crisis detection was assessed due to availability of clinician-reviewed labels (i.e., only crisis vs. non-crisis labels were generated by clinicians as part of this study).

For the comparative analysis, the VBHSF was compared against benchmark models on simulated (n=1,800) and external (n=794) datasets. For overall crisis detection, differences in the primary outcomes of sensitivity and specificity were assessed using an omnibus Cochran's Q test, followed by pairwise McNemar's tests with a Bonferroni correction ($\alpha = 0.017$) for multiple comparisons. Operational utility was evaluated by projecting the positive predictive value (PPV) across a range of real-world prevalence rates.

For the per-category analysis on the Verily simulated dataset, our model's hierarchical predictions were evaluated as a "flat" system to ensure identical denominators across all models. Differences in metrics were assessed with the same statistical approach. Due to incompatible crisis categories across models, the benchmark models' overall crisis flag was evaluated against each specific ground truth.

## Results

**Dataset Descriptives**

The NVIDIA dataset had more characters, words, sentences, unique tokens, and total tokens than the Verily dataset, but less lexical diversity. Both datasets shared 60% of their top 10 most frequent words, with similar overall proportional representation (see Table 1).

**VBHSF Evaluation.**

On the Verily dataset, the VBHSF had a sensitivity of 0.99 (95% CI: 0.981-0.995) and specificity of 0.992 (95% CI: 0.984-0.996) for crisis vs non-crisis (see Figure 1, Table 2). The multi-label classifier, showed a macro-averaged F1-score of 0.939 (95% CI: 0.927-0.951), macro-averaged sensitivity of 0.957 (95% CI: 0.937-0.978), and PPV of 0.923 (95% CI: 0.894-0.952). Per-category sensitivity ranged from 0.917-0.992, and specificity was at or above 0.978 for all crisis types (Table 3). Among the 898 messages flagged as "Crisis", 88 messages (9.8%) contained at least one category misclassification, resulting in 125 individual category (false positive or false negative) errors (see Supplementary Materials). On the NVIDIA dataset, the VBHSF had sensitivity of 0.982 (95% CI: 0.964-0.991) and specificity of 0.859 (95% CI: 0.821-0.889).

**Comparative Evaluation**

In the comparative evaluation on the Verily dataset (n=1,800), the VBHSF demonstrated the highest overall performance, with a sensitivity of 0.990 (95% CI: 0.981-0.995) and a specificity of 0.992 (95% CI: 0.984-0.996) (see Figure 1, Table 2). OpenAI model achieved the highest specificity of 0.999 (95% CI: 0.994-1.000) with a sensitivity of 0.419 (95% CI:

0.387-0.451), while NVIDIA's model had sensitivity of 0.759 (95% CI: 0.730-0.786) and specificity of 0.756 (95% CI: 0.726-0.783). A Cochran's Q test revealed a significant difference among the models for both sensitivity ($Q(2) = 751.12$, $p < 0.001$) and specificity ($Q(2) = 418.68$, $p < 0.001$). Pairwise McNemar's tests with Bonferroni correction revealed the VBHSF's sensitivity was significantly higher than the NVIDIA and OpenAI guardrails ($p < 0.001$). The VBHSF's specificity was significantly higher than the NVIDIA ($p < 0.001$), but was not significantly different from the OpenAI guardrail ($p = 0.094$).

To assess generalizability, all models were evaluated on the NVIDIA dataset. The VBHSF showed the highest sensitivity (0.982, 95% CI: 0.964-0.991), which was significantly higher than both OpenAI (0.882, $p < 0.001$) and NVIDIA (0.907, $p < 0.001$) guardrails. No significant differences in specificity were observed among the guardrails ($Q(2) = 4.28$, $p = 0.118$). Table 2 shows a comprehensive summary of all performance metrics for each model across datasets.

In the comparative evaluation of specific mental health crisis detection, VBHSF sensitivity ranged from 0.88-0.992 and accuracy ranged from 0.987-0.995. The OpenAI sensitivity ranged from 0.097-0.916 and accuracy ranged from 0.739-0.845. NVIDIA sensitivity ranged from 0.503-1.000 and accuracy ranged from 0.500-0.565. Detailed performance metrics for all models across all categories are presented in Table 2.

To assess operational utility, PPV was projected across a range of plausible low-prevalence crisis rates for both datasets (see Supplementary Tables 1 and 2). On the Verily dataset with an assumed 2% prevalence rate the PPV was 0.895 (95% CI: 0.717-0.989) for OpenAI, 0.716 (95% CI: 0.576-0.822) for VBHSF, and 0.060 (95% CI: 0.041-0.085) for NVIDIA (see Supplementary Materials).

**Discussion**

This study demonstrated the feasibility of building and evaluating a fit-for-purpose safety filter for text-based LLM applications. The VBHSF's exhibited strong performance in accurately detecting mental health crises and classifying crises into eight specific categories linked to adverse outcomes. Evaluation used two clinician-labeled datasets: Verily Mental Health Crisis Dataset v1.0 of 1,800 simulated messages reflecting real-world texting behaviors, and the refined NVIDIA Aegis AI Content Safety Dataset 2.0. VBHSF significantly outperformed state-of-the-art general purpose safety guardrails in overall sensitivity for crisis detection.

The VBHSF had high sensitivity (0.990) and specificity (0.992) for detecting mental health crises, indicating that this safety filter rarely missed true crisis messages, while maintaining a very low false-positive rate, an important balance for clinical safety systems where both over- and under-flagging can carry consequences. When examined at the crisis category level, the VBHSF demonstrated high macro-averaged F1-score of 0.939 with per-category sensitivity ranging from 0.917-0.992, and specificity at or above 0.978 for all crisis types, suggesting robust performance across all crisis categories. When evaluated against the NVIDIA Aegis AI Content Safety Dataset 2.0, VBHSF performance remained highly sensitive (0.982) and accuracy (0.921) with reduced specificity (0.859). This pattern suggests that the model remains highly effective in minimizing false negatives, critical for safety applications, while showing some increase in false-positive detections in a novel sample, a trade-off that may be acceptable in high-risk healthcare contexts.

Evaluation of two general-purpose safety guardrail models for mental health crisis detection contextualized the performance of the VBHSF across datasets. On the Verily dataset, the NVIDIA model showed balanced sensitivity (0.759) and specificity (0.756), while the

OpenAI model prioritized specificity (0.999) at the expense of sensitivity (0.419). On the NVIDIA dataset, NVIDIA achieved higher sensitivity (0.907) and specificity (0.886), with OpenAI displaying similarly balanced performance on sensitivity (0.882) and specificity (0.899). Across both datasets, the VBHSF significantly outperformed these guardrails on key performance metrics, achieving higher sensitivity in all cases and higher specificity relative to NVIDIA, but not to OpenAI. These findings suggest the VBHSF offers superior crisis detection, while maintaining comparable or lower false-positive rates. When examining the ability to identify specific mental health crisis categories, the VBHSF demonstrated high and consistent performance, with sensitivity ranging from 0.880-0.992 and accuracy from 0.987-0.995. These results indicate that VBHSF reliably detects true crisis cases while minimizing classification errors, an essential property for deployment in safety-critical settings designed to connect individuals with personalized crisis resources. In contrast, OpenAI exhibited considerably lower and more variable sensitivity (0.097-0.916) and accuracy (0.739-0.845), suggesting potential under-identification of crisis categories. NVIDIA displayed wider sensitivity ranging from 0.503-1.000 combined with low accuracy of 0.500-0.565 highlighting inconsistent detection, potentially leading to a high rate of false positives or negatives depending on crisis category. Because the Nvidia and OpenAI guardrails did not cover all of the same risk categories as the VBHSF, this variability in performance across some categories was expected. .

Overall, the VBHSF offers balanced sensitivity and specificity, minimizing false positives (7) and negatives (9) compared to OpenAI, which, despite having fewer false positives (1), has substantially higher false negatives (523), and NVIDIA, which produced high false positives (220) and negatives (217). This indicates that the VBHSF is well-suited for deployment in scenarios where avoiding missed crises, reducing alert fatigue, and pushing personalized crisis resources are critical, such as clinical triage systems, integrated behavioral health platforms, and large-scale digital health interventions requiring scalable and safety focused applications. However, projected PPV in a low-prevalence environment provided a more nuanced picture of real-world performance. While OpenAI's very high specificity (0.999) resulted in a slightly higher projected PPV than VBHSF on the Verily dataset, this advantage did not hold on the NVIDIA dataset, where both experienced a drop in specificity, resulting in uniformly low and statistically indistinguishable performance. These results highlight a crucial consideration for deployment in low-prevalence settings: even best-performing models will inevitably produce a substantial number of false alarms, precluding fully autonomous use without unnecessarily burdening users who are not in crisis. Rather than replacing human judgement, the VBHSF's high sensitivity makes it most valuable as a screening tool, with human-in-the-loop oversight essential to adjudicate alerts, filter false positives to prevent alert fatigue, and ensure timely and appropriate crisis resources for those at risk.

**Limitation and Future Directions**

Despite strengths like a diverse, clinician-labeled dataset of mental health messages, and generalizing performance to an NVIDIA dataset, the study has limitations. Messages were English-only, which may restrict applicability across languages, though LLMs often transfer well across languages. Future studies should simulate multi-language data. All messages were single-turn, and safety guardrails are known to fail more frequently in multi-turn conversations, suggesting performance may drop[8]. The NVIDIA dataset[29] included messages used to train their guardrail, so the reported performance may not generalize and should be interpreted with caution as a possible source of methodological bias. Finally, all messages were simulated, not real user messages.

This initial work establishes a framework for identifying psychiatric crises in user prompts. The industry urgently needs a standardized assessment framework for safety concerns, coupled with representative and clinician annotated datasets with real-world multi-turn conversations and be updated to capture evolving slang and coded language used in crises. Standardized protocols for red-teaming, adversarial testing, and post-deployment monitoring are needed to ensure safety as technology and contexts evolve. These steps will move the field towards a shared infrastructure for improving mental health safety guardrails, ensuring LLMs are reliable, clinically informed, and protective of users in crisis.

**Conclusion**

This work presents an advancement in digital mental health safety. We successfully developed the VBHSF, a fit-for-purpose mental health safety filter that detects and categorizes subtle crisis signals. VBHSF showed high, generalized performance on rigorously annotated, internal and external datasets. Its balanced sensitivity and specificity indicate potential suitability for healthcare settings that must manage both crisis detection and alert fatigue, though confirmation through future red-teaming, adversarial testing, and validation on patient data will be essential before deployment in healthcare contexts, such as clinical triage, integrated behavioral health platforms, or large-scale digital health conversational agents.

# Disclosures/Acknowledgements

**Declaration of conflicts of interest:** BWN, CW, MTS, SS, AR, JL, EY, and AT report employment and equity ownership in Verily Life Sciences.

**Author contributions:**

Study concept and design: BWN

Data collection: BWN, AR, JT

Data analysis and interpretation: CW, BWN, JT, AT, MS, SS, JL

Draft writing and review: BWN wrote the initial draft and all authors reviewed

Draft approval for submission: BWN, JT, AT

**Funding**:

This study was funded by Verily Life Sciences

**Prior disclosure of these data:**

arxiv

**Data sharing statement:**

Data from this study are available upon researcher request.

**Acknowledgements**:

Authors wish to acknowledge NVIDIA for providing open access to the NVIDIA Aegis AI Content Safety Dataset 2.0.

**Tables**

Table 1. Descriptive Statistics of the Datasets

| **Message Length and Structure** | | | | |
|---|---|---|---|---|
| **Dataset** | **Statistic** | **Characters per message** | **Words per message** | **Sentences per message** |
| Verily Mental Health Crisis Dataset v1.0 | Mean (SD) | 96.70 (84.50) | 18.07 (15.14) | 1.50 (0.82) |
| | Median | 51.50 | 12.00 | 1.00 |
| | Min-Max | 21 - 511 | 2 - 99 | 1 - 6 |
| NVIDIA Aegis AI Content Safety Dataset 2.0 | Mean (SD) | 364.46 (849.37) | 70.67 (163.07) | 4.97 (10.48) |
| | Median | 65.00 | 13.00 | 1.00 |
| | Min-Max | 2 - 9,940 | 1 - 1,963 | 1 - 135 |

| **Lexical Properties** | | | |
|---|---|---|---|
| **Dataset** | **Unique Tokens (Vocabulary Size)** | **Total Tokens (Total Words)** | **Type-Token Ratio** |
| Verily Mental Health Crisis Dataset v1.0 | 4,695 | 32,527 | 0.144 |
| NVIDIA Aegis AI Content Safety Dataset 2.0 | 7,526 | 56,115 | 0.134 |

| **Top 10 Most Frequent Words** | | | |
|---|---|---|---|
| Verily Mental Health Crisis Dataset v1.0 | | NVIDIA Aegis AI Content Safety Dataset 2.0 | |
| I | 1706 (5.245%) | I | 3308 (5.895%) |
| My | 1394 (4.286%) | To | 2003 (3.569%) |
| To | 814 (2.503%) | And | 1562 (2.783%) |
| A | 544 (1.672%) | The | 1265 (2.254%) |
| I'm | 524 (1.611%) | A | 1167 (2.080%) |

| Me  | 480 (1.476)   | My   | 1112 (1.982%) |
|-----|---------------|------|---------------|
| The | 437 (1.334%)  | Of   | 855 (1.524%)  |
| Can | 434 (1.334%)  | It   | 609 (1.085%)  |
| And | 423 (1.301%)  | That | 592 (1.055%)  |
| For | 356 (1.094%)  | In   | 590 (1.051%)  |

Table 2. Performance of Safety Guardrail Models on Overall Crisis Detection for Simulated and NVIDIA datasets

| Performance Metric | Verily behavioral health safety filter | | OpenAI Omni | | NVIDIA NeMo | |
|---|---|---|---|---|---|---|
| | Verily Dataset | NVIDIA Dataset | Verily Dataset | NVIDIA Dataset | Verily Dataset | NVIDIA Dataset |
| Accuracy[1] | 0.991 (0.986-0.995) | 0.921 (0.900-0.937) | 0.709 (0.687-0.729) | 0.890 (0.867, 0.910) | 0.757 (0.737, 0.776) | 0.897 (0.874, 0.916) |
| Sensitivity[1,3] | 0.990 (0.981-0.995) | 0.982 (0.964-0.991) | 0.419 (0.387-0.451) | 0.882 (0.846, 0.910) | 0.759 (0.730, 0.786) | 0.907 (0.874, 0.932) |
| Specificity[2] | 0.992 (0.984-0.996) | 0.859 (0.821-0.889) | 0.999 (0.994-1.000) | 0.899 (0.865, 0.925) | 0.756 (0.726, 0.783) | 0.886 (0.851, 0.914) |
| PPV | 0.992 (0.984-0.996) | 0.875 (0.841-0.902) | 0.997 (0.985-1.000) | 0.898 (0.864, 0.924) | 0.756 (0.727, 0.783) | 0.889 (0.855, 0.916) |
| NPV | 0.990 (0.981-0.995) | 0.980 (0.959, 0.990) | 0.632 (0.607-0.657) | 0.883 (0.848, 0.911) | 0.758 (0.729, 0.785) | 0.905 (0.871, 0.930) |
| False Positive | 7 | 56 | 1 | 40 | 220 | 45 |
| False Negative | 9 | 7 | 523 | 47 | 217 | 37 |

All values are proportions with 95% confidence intervals in parentheses. All pairwise comparisons are Bonferroni-corrected McNemar's tests following a significant omnibus Cochran's Q test. Statistical comparisons for PPV and NPV were not performed; their denominators vary by model, which violates the assumptions of paired statistical tests.
PPV = Positive Predictive Value; NPV = Negative Predictive Value.
***Verily Mental Health Crisis Dataset v1.0 (n=1,800) statistics:***
[1]Verily > OpenAI and NVIDIA (p<0.0001 for both).
[2]Verily > NVIDIA (p<0.0001).
***NVIDIA Aegis AI Content Safety Dataset v2.0 (n=794) statistics:***
[3]Verily > OpenAI and NVIDIA (p<0.0001 for both).

Table 3. Performance of Safety Guardrail Models by Crisis Type Using Verily Mental Health Crisis Dataset v1.0 (n=1,800)

| Crisis Type | Accuracy (95% CI) | Sensitivity (95% CI) | Specificity (95% CI) | PPV (95% CI) | NPV (95% CI) | False Positive | False Negative |
|---|---|---|---|---|---|---|---|
| **Verily behavioral health safety filter** | | | | | | | |
| All | 0.991 (0.986, 0.995) | 0.990 (0.981, 0.995) | 0.992 (0.984, 0.996) | 0.992 (0.984, 0.996) | 0.990 (0.981, 0.995) | 7 | 9 |
| Abuse | 0.989 (0.984, 0.993) | 0.985 (0.947, 0.996) | 0.990 (0.984, 0.994) | 0.886 (0.825, 0.928) | 0.999 (0.996, 1.000) | 17 | 2 |
| Eating Disorder Behavior | 0.990 (0.984, 0.994) | 0.880 (0.811, 0.926) | 0.998 (0.995, 0.999) | 0.973 (0.925, 0.991) | 0.991 (0.985, 0.995) | 3 | 15 |
| Neglect | 0.990 (0.984, 0.994) | 0.947 (0.889, 0.975) | 0.993 (0.988, 0.996) | 0.899 (0.832, 0.941) | 0.996 (0.992, 0.998) | 12 | 6 |
| Psychosis | 0.993 (0.988, 0.996) | 0.962 (0.913, 0.983) | 0.995 (0.991, 0.998) | 0.940 (0.886, 0.969) | 0.997 (0.993, 0.999) | 8 | 5 |
| Self Harm | 0.992 (0.987, 0.995) | 0.939 (0.884, 0.969) | 0.996 (0.992, 0.998) | 0.953 (0.902, 0.979) | 0.995 (0.991, 0.998) | 6 | 8 |
| Substance Misuse | 0.989 (0.984, 0.993) | 0.942 (0.889, 0.970) | 0.993 (0.988, 0.996) | 0.921 (0.865, 0.956) | 0.995 (0.991, 0.998) | 11 | 8 |
| Suicide | 0.987 (0.980, 0.991) | 0.944 (0.890, 0.973) | 0.990 (0.984, 0.994) | 0.875 (0.809, 0.920) | 0.996 (0.991, 0.998) | 17 | 7 |
| Violence Towards Others | 0.995 (0.991, 0.997) | 0.992 (0.954, 0.999) | 0.995 (0.991, 0.998) | 0.937 (0.880, 0.967) | 0.999 (0.997, 1.000) | 8 | 1 |
| **OpenAI Omni Moderation** | | | | | | | |
| All | 0.709 (0.687, 0.729) | 0.419 (0.387, 0.451) | 0.999 (0.994, 1.000) | 0.997 (0.985, 1.000) | 0.632 (0.607, 0.657) | 1 | 523 |
| Abuse[1] | 0.789 (0.769, 0.807) | 0.493 (0.409, 0.576) | 0.813 (0.793, 0.831) | 0.175 (0.140, 0.216) | 0.952 (0.940, 0.962) | 68 | 312 |
| Eating Disorder Behavior[1] | 0.758 (0.738, 0.778) | 0.272 (0.202, 0.356) | 0.795 (0.775, 0.813) | 0.090 (0.065, 0.123) | 0.936 (0.922, 0.948) | 91 | 344 |
| Neglect | 0.739 (0.719, 0.759) | 0.097 (0.055, 0.166) | 0.782 (0.762, 0.801) | 0.029 (0.016, 0.051) | 0.928 (0.914, 0.941) | 102 | 367 |
| Psychosis | 0.748 (0.727, 0.767) | 0.208 (0.147, 0.285) | 0.790 (0.770, 0.809) | 0.071 (0.050, 0.102) | 0.928 (0.913, 0.940) | 103 | 351 |
| Self Harm[2] | 0.819 (0.801, 0.837) | 0.702 (0.619, 0.774) | 0.829 (0.810, 0.846) | 0.243 (0.203, 0.289) | 0.973 (0.963, 0.980) | 39 | 286 |
| Substance Misuse[1] | 0.767 (0.747, 0.786) | 0.350 (0.276, 0.433) | 0.802 (0.782, 0.820) | 0.127 (0.097, 0.164) | 0.937 (0.924, 0.949) | 89 | 330 |
| Suicide[2] | 0.814 (0.796, 0.832) | 0.675 (0.589, 0.750) | 0.825 (0.806, 0.842) | 0.225 (0.186, 0.270) | 0.971 (0.961, 0.979) | 41 | 293 |
| Violence Towards Others[2] | 0.845 (0.828, 0.861) | 0.916 (0.852, 0.954) | 0.840 (0.822, 0.857) | 0.288 (0.245, 0.336) | 0.993 (0.987, 0.996) | 10 | 269 |
| **NVIDIA NeMo** | | | | | | | |
| All | 0.757 (0.737, 0.776) | 0.759 (0.730, 0.786) | 0.756 (0.726, 0.783) | 0.756 (0.727, 0.783) | 0.758 (0.729, 0.785) | 220 | 217 |
| Abuse[1] | 0.565 (0.542, 0.588) | 0.948 (0.896, 0.974) | 0.534 (0.510, 0.558) | 0.141 (0.119, 0.165) | 0.992 (0.984, 0.996) | 7 | 776 |
| Eating Disorder Behavior | 0.510 (0.487, 0.533) | 0.584 (0.496, 0.667) | 0.504 (0.481, 0.528) | 0.081 (0.065, 0.100) | 0.942 (0.925, 0.956) | 52 | 830 |
| Neglect | 0.500 (0.477, 0.523) | 0.513 (0.422, 0.603) | 0.499 (0.475, 0.523) | 0.064 (0.050, 0.082) | 0.939 (0.921, 0.953) | 55 | 845 |
| Psychosis | 0.525 (0.502, 0.548) | 0.685 (0.600, 0.758) | 0.513 (0.489, 0.536) | 0.099 (0.081, 0.120) | 0.954 (0.939, 0.966) | 41 | 814 |
| Self Harm[2] | 0.548 (0.525, 0.571) | 0.840 (0.767, 0.893) | 0.525 (0.501, 0.549) | 0.122 (0.102, 0.145) | 0.977 (0.964, 0.985) | 21 | 793 |
| Substance Misuse[2] | 0.561 (0.538, 0.584) | 0.912 (0.853, 0.949) | 0.532 (0.508, 0.556) | 0.138 (0.117, 0.162) | 0.987 (0.977, 0.992) | 12 | 778 |
| Suicide[2] | 0.535 (0.512, 0.558) | 0.762 (0.680, 0.828) | 0.518 (0.494, 0.542) | 0.106 (0.088, 0.128) | 0.967 (0.953, 0.976) | 30 | 807 |
| Violence Towards Others[2] | 0.564 (0.541, 0.587) | 1.000 (0.969, 1.000) | 0.534 (0.510, 0.557) | 0.132 (0.111, 0.155) | 1.000 (0.996, 1.000) | 0 | 784 |

Note: [1] indicates the crisis categories that partially overlap with the intended use of the guardrail; [2] indicates the crisis categories that directly overlaps with intended use of each guardrail.

Table 4. Internal Evaluation of the Verily Mental Health Guardrails Hierarchical System: Per-Category Performance (n=898)

| Crisis Type | Accuracy (95% CI) | Sensitivity (95% CI) | Specificity (95% CI) | PPV (95% CI) | NPV (95% CI) | False Positive | False Negative |
|---|---|---|---|---|---|---|---|
| Abuse | 0.979 (0.967, 0.986) | 0.985 (0.947, 0.996) | 0.978 (0.965, 0.986) | 0.886 (0.825, 0.928) | 0.997 (0.990, 0.999) | 17 | 2 |
| Eating Disorder Behaviors | 0.986 (0.975, 0.992) | 0.917 (0.853, 0.954) | 0.996 (0.989, 0.999) | 0.973 (0.925, 0.991) | 0.987 (0.977, 0.993) | 3 | 10 |
| Neglect | 0.981 (0.970, 0.988) | 0.955 (0.900, 0.981) | 0.985 (0.974, 0.991) | 0.899 (0.832, 0.941) | 0.994 (0.985, 0.997) | 12 | 5 |
| Psychosis | 0.986 (0.975, 0.992) | 0.962 (0.913, 0.983) | 0.99 (0.980, 0.995) | 0.94 (0.886, 0.969) | 0.993 (0.985, 0.997) | 8 | 5 |
| Self-Harm | 0.984 (0.974, 0.991) | 0.939 (0.884, 0.969) | 0.992 (0.983, 0.996) | 0.953 (0.902, 0.979) | 0.99 (0.980, 0.995) | 6 | 8 |
| Substance Misuse | 0.979 (0.967, 0.986) | 0.942 (0.889, 0.970) | 0.986 (0.974, 0.992) | 0.921 (0.865, 0.956) | 0.989 (0.979, 0.995) | 11 | 8 |
| Suicide | 0.977 (0.965, 0.985) | 0.967 (0.919, 0.987) | 0.978 (0.965, 0.986) | 0.875 (0.809, 0.920) | 0.995 (0.987, 0.998) | 17 | 4 |
| Violence Towards Others | 0.99 (0.981, 0.995) | 0.992 (0.954, 0.999) | 0.99 (0.980, 0.995) | 0.937 (0.880, 0.967) | 0.999 (0.993, 1.000) | 8 | 1 |

# Figures

a.

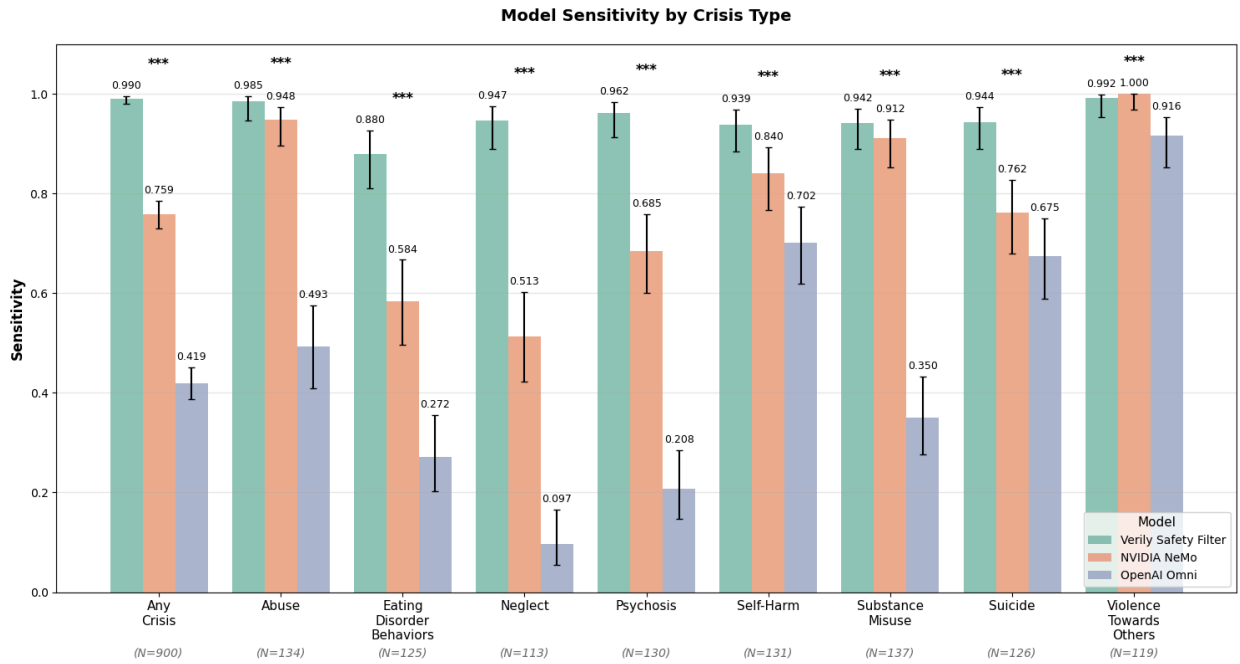

b.

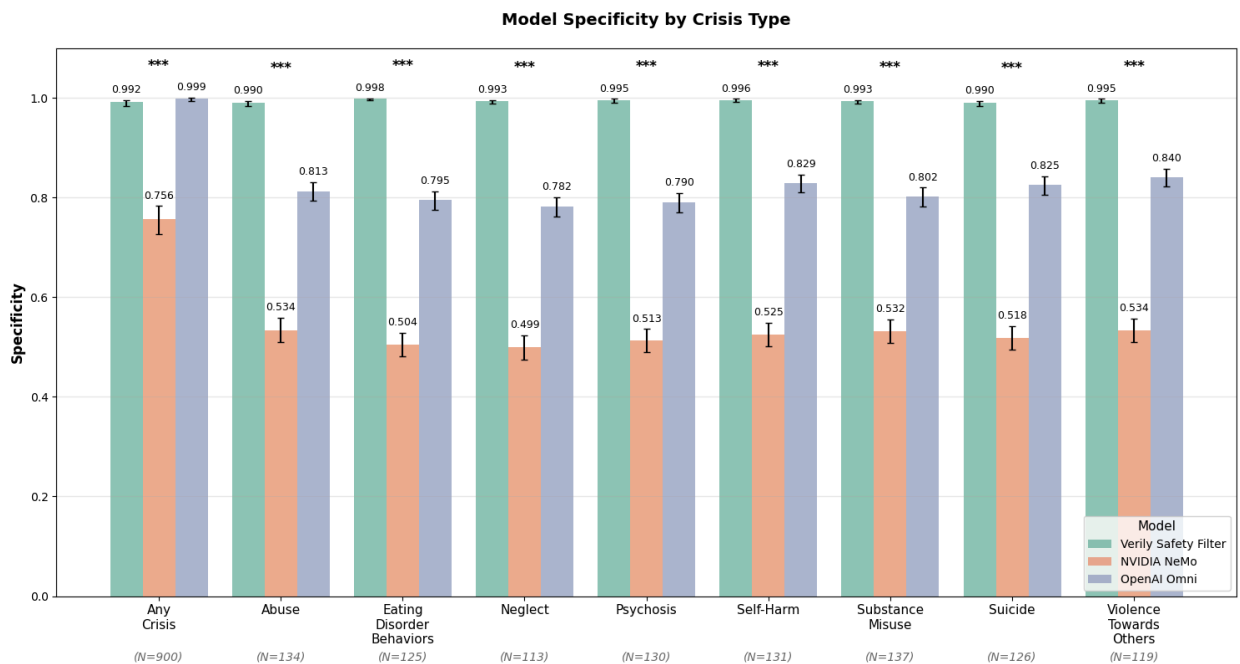

Figure 1. Sensitivity (a) and Specificity (b) for Any Crisis and Crisis Types Across Three Guardrail Models on the Verily Dataset.
Note: *p<0.0167 (Bonferroni correction); **p < 0.01; ***p < 0.001. P-values for differences between models in performance metrics are based on Cochran's Q tests.

**Supplemental Material**

Table S1. Projected PPV for Different Crisis Message Prevalence Rates (Verily Mental Health Crisis Dataset v1.0)

| Model | Sensitivity | Specificity | 0.5% Prevalence | 1% Prevalence | 2% Prevalence | 5% Prevalence |
|---|---|---|---|---|---|---|
| | | | PPV (95% CI) | PPV (95% CI) | PPV (95% CI) | PPV (95% CI) |
| NVIDIA | 0.759 (0.730, 0.786) | 0.756 (0.726, 0.783) | 0.015 (0.006, 0.029) | 0.03 (0.017, 0.049) | 0.06 (0.041, 0.085) | 0.141 (0.112, 0.174) |
| Openai | 0.419 (0.387, 0.451) | 0.999 (0.994, 1.000) | 0.678 (0.301, 0.954) | 0.809 (0.529, 0.978) | 0.895 (0.717, 0.989) | 0.957 (0.865, 0.995) |
| Verily | 0.990 (0.981, 0.995) | 0.992 (0.984, 0.996) | 0.383 (0.197, 0.570) | 0.556 (0.378, 0.708) | 0.716 (0.576, 0.822) | 0.867 (0.794, 0.924) |

Table S2. Projected PPV for Different Crisis Message Prevalence Rates (NVIDIA Aegis AI Content Safety Dataset 2.0)

| Model | Sensitivity | Specificity | 0.5% Prevalence | 1% Prevalence | 2% Prevalence | 5% Prevalence |
|---|---|---|---|---|---|---|
| | | | PPV (95% CI) | PPV (95% CI) | PPV (95% CI) | PPV (95% CI) |
| NVIDIA | 0.907 (0.874, 0.932) | 0.886 (0.851, 0.914) | 0.038 (0.019, 0.073) | 0.074 (0.045, 0.115) | 0.14 (0.099, 0.187) | 0.295 (0.244, 0.351) |
| OpenAI | 0.882 (0.846, 0.910) | 0.899 (0.865, 0.925) | 0.042 (0.018, 0.075) | 0.081 (0.047, 0.124) | 0.151 (0.107, 0.203) | 0.315 (0.260, 0.375) |
| Verily | 0.982 (0.964, 0.991) | 0.859 (0.821, 0.889) | 0.034 (0.016, 0.060) | 0.066 (0.040, 0.099) | 0.124 (0.090, 0.167) | 0.268 (0.223, 0.318) |

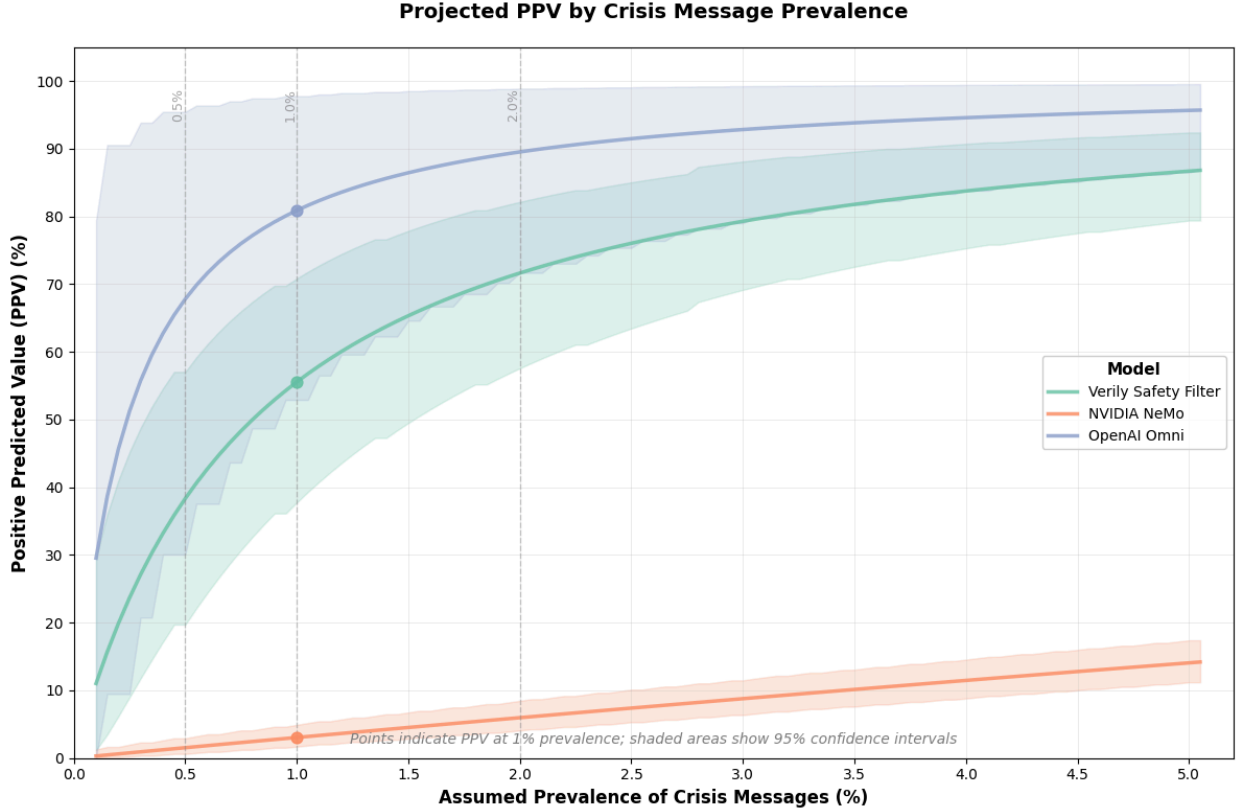

Figure S1. Projected Positive Predicted Values by Crisis Message Prevalence Across Three Guardrail Models on Verily Mental Health Crisis Dataset v1.0

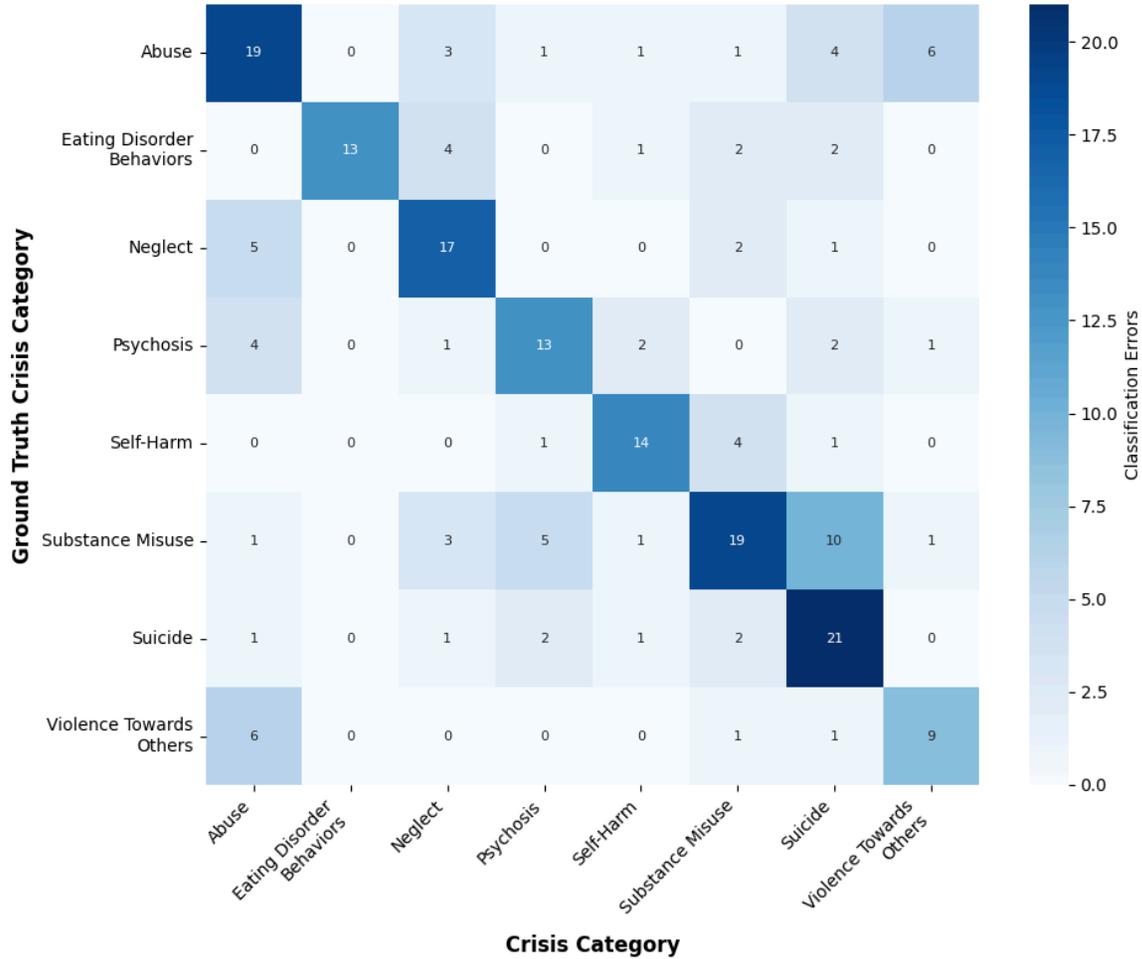

Figure S2. Misclassification Matrix for Verily Safety Filter Stage 2 Multi-Label Classifier (n=898)

Note: Rows represent ground truth crisis categories. Diagonal values show direct classification errors for each ground truth category (82 false positives + 43 false negatives). Off-diagonal values show cross-category confusion patterns where the row category was truly present but the column category was incorrectly predicted. Only diagonal errors (125 total) contribute to overall error statistics.